%% file: main.tex
\documentclass{article}

\usepackage[final]{corl_2022} %

\title{COACH: Cooperative Robot Teaching}

\usepackage{bbm}
\usepackage{graphicx}
\usepackage{cellspace}
\usepackage{caption}
\usepackage{algorithm}
\usepackage{algpseudocode}
\usepackage[usenames,dvipsnames]{xcolor}

\newcommand\cincludegraphics[2][]{\raisebox{-0.3\height}{\includegraphics[#1]{#2}}}
\usepackage{wrapfig}
\input{var}

\input{macros}

\author{
 Cunjun Yu$^{1}$
  \quad
 Yiqing Xu$^{1}$
  \quad
 Linfeng Li$^{1}$
  \quad
  David Hsu$^{1,2}$  \vspace{0.25cm}\\
  $^1$School of Computing  \\
  $^2$Smart Systems Institute\\
  National University of Singapore\\
}

\begin{document}
\maketitle
\input{sections/0_abstract.tex}
\input{sections/1_intro.tex}

\input{sections/2_related_work.tex}

\input{sections/3_formulation.tex}

\input{sections/4_method.tex}

\input{sections/5_experiment.tex}

\input{sections/6_limitation.tex}

\input{sections/7_conclusion.tex}
\bibliography{ref}  %
\end{document}

%% file: var.tex
\usepackage{amsmath,amssymb}
\newenvironment{tightlist}%
{\begin{list}{$\bullet$}{%
    \setlength{\topsep}{0in}
    \setlength{\partopsep}{0in}
    \setlength{\itemsep}{0.0in}
    \setlength{\parsep}{0in}
    \setlength{\leftmargin}{3.5em}
    \setlength{\rightmargin}{0in}
    \setlength{\itemindent}{0in}
}
}%
{\end{list}}

\newcommand{\dataset}{\mathbb{X}}
\newcommand{\sss}{\scriptscriptstyle}
\newcommand{\ballmaze}{Cooperative Ball Maze}
\newcommand{\overcooked}{Overcooked-AI}
\newcommand{\TimeInterval}{\ensuremath{\Delta t}}

\newcommand{\ttt}{the target task}
\newcommand{\eg}{e.g.}

\newcommand{\student}{S}
\newcommand{\agenta}{\ensuremath{1}}
\newcommand{\agentb}{\ensuremath{2}}
\newcommand{\optimalpolicy}{\ensuremath{\phi^*}}

\newcommand{\Bpolicy}{\ensuremath{\pi^\agentb{}}}
\newcommand{\Apolicy}{\ensuremath{\pi^\agenta{}}}
\newcommand{\Tpolicy}{\ensuremath{\bar{\pi}}}
\newcommand{\Spolicy}{\ensuremath{\phi}}

\newcommand{\statespace}{\ensuremath{S}}
\newcommand{\state}{\ensuremath{s}}

\newcommand{\Tdots}{\ensuremath{T}}
\newcommand{\Rdots}{\ensuremath{R}}

\newcommand{\Cdots}{\ensuremath{C}}

\newcommand{\studentsuper}{\ensuremath{\mathrm{\student}}}

\newcommand{\Aaction}{\ensuremath{a^\agenta}}
\newcommand{\Baction}{\ensuremath{a^\agentb}}
\newcommand{\Aactionspace}{\ensuremath{A^\agenta}}
\newcommand{\Bactionspace}{\ensuremath{A^\agentb}}

\newcommand{\Saction}{\ensuremath{a^{\sss \studentsuper}}}

\newcommand{\Supdate}{\ensuremath{U}}
\newcommand{\PomdpStateSpace}{\ensuremath{\bar{S}}}
\newcommand{\PomdpState}{\ensuremath{\bar{s}}}
\newcommand{\PomdpActionSpace}{\ensuremath{\bar{A}}}
\newcommand{\PomdpAction}{\ensuremath{\bar{a}}}

\newcommand{\PomdpTransition}{\ensuremath{\bar{T}}}

\newcommand{\PomdpRewardFunction}{\ensuremath{\bar{R}}}
\newcommand{\PomdpRewardFunctionDots}{\ensuremath{\bar{R}}}

\newcommand{\PomdpDistanceMeasure}{\ensuremath{D}}
\newcommand{\PomdpDistance}{\ensuremath{D}}
\newcommand{\PomdpObservation}{\ensuremath{\bar{O}}}
\newcommand{\SmallPomdpObservation}{\ensuremath{\bar{o}}}
\newcommand{\PomdpObservationFunction}{\ensuremath{\bar{{Z}}}}
\newcommand{\PomdpGamma}{\ensuremath{\bar{\gamma}}}

%% file: macros.tex
\newtheorem{defn}{Definition}

\usepackage{ifthen}
\newboolean{BoolComment}
\setboolean{BoolComment}{true}

 \usepackage[dvipsnames]{xcolor}

\newcounter{ccmt}
\newcommand{\comment}[4]{
  \ifthenelse{\boolean{BoolComment}}
  {\addtocounter{ccmt}{1}
    \textcolor{#2}{#3}
    \marginpar{\tiny\noindent{\raggedright{{\colorbox{#2}{\sffamily\textcolor{white}{#1
                [\arabic{ccmt}]}}}}} \textcolor{#2}{#4} \par}}
  {#3}}

\newlength{\minitextwidth}
\usepackage{amsmath,amssymb}

\newcommand{\captionfonts}{\small}
\makeatletter                   %
\long\def\@makecaption#1#2{%
  \vskip\abovecaptionskip
  \sbox\@tempboxa{{\captionfonts #1. #2}}%
  \ifdim \wd\@tempboxa >\hsize
    {\captionfonts #1. #2\par}
  \else
    \hbox to\hsize{\hfil\box\@tempboxa\hfil}%
  \fi
  \vskip\belowcaptionskip}
\makeatother                    %

\newcommand{\qed}{\mbox{}\hspace*{\fill}\nolinebreak\mbox{$\Box$}\smallskip}

\newcommand{\argmax}{\mathop{\mathrm{arg\,max}}}

%% file: sections/0_abstract.tex
\begin{abstract}
  Knowledge and skills can transfer from human teachers to human
  students. However, such direct transfer is often not scalable for
  \textit{physical} tasks, as they require one-to-one interaction, and human
  teachers are not available in sufficient numbers.  Machine learning enables
  robots to become experts and play the role of teachers to help in this
  situation.  In this work, we formalize \textit{cooperative robot teaching} 
  as a Markov game, consisting of four key elements: the target task, the
  student model, the teacher model, and the interactive
  teaching-learning process.  Under a moderate assumption, the Markov game
  reduces to a partially observable Markov decision process, with an efficient
  approximate solution. We illustrate our approach on two cooperative tasks, one in a
  simulated video game and one with a real robot.
\end{abstract}

%% file: sections/1_intro.tex
\section{Introduction}  

How do we teach humans to re-orientate a table jointly or play tennis? Humans
often learn by practicing the skills with teachers or
partners~\cite{Gillies2016CooperativeLR, Ross1995DifferentiatingCL,
  Omidshafiei2019}.  This mode of learning is, however, difficult to scale up,
as it requires one-to-one interaction and there are not sufficient human
teachers~\cite{garcia2019teacher}.  With advances in machine learning, 
robots can not only master complex tasks~\cite{alphago, helicopter,ikea} but
also collaborate with humans and adapt to human behaviors~\cite{crosstraining,
  trust, cirl}.  In this work, we aim to create \emph{robot teachers} for
physical tasks, thus scaling up teaching and providing learning opportunities
to many even when human teachers are not available.

Specifically, we propose \emph{Cooperative rObot teACHing} (COACH),
a robot teaching framework to teach humans cooperative skills for two-player
physical tasks through interaction. We assume the robot teacher has full knowledge of the task, specifically,  a set of policies to execute the task. The objective is to teach the student a policy as fast as possible. See Fig.~\ref{fig:overview} for an illustration.
COACH treats the teaching task as a two-player Markov game for a \textit{target task}. One player is the
robot teacher, and the other is the human student. Under a suitable student
learning model,  COACH transforms the game into a \emph{partially observable Markov
decision process}  (POMDP).  The POMDP solution enables the robot teacher to adapt to
the different behaviors, according to the history of interactions.

One key challenge of COACH is to represent the student's knowledge of the target skills and 
learning behaviors. First, we leverage \textit{item response theory} (IRT), a well-established framework for
educational assessment~\cite{irt}. IRT provides simplified parametric models that
capture the student's knowledge level with respect to the task difficulty in a small number of parameters.
COACH treats these parameters as latent variables in the teaching POMDP and learns them from human-robot interaction data by solving the POMDP. 
 Next, to teach complex skills, we 
 draw insights from student-centered
learning~\cite{jones2007student} and human-robot
cross-training~\cite{Nikolaidis2013}.
We decompose a complex target skill into a  set of sub-skills, based on the student's potential roles in the target task.
With this compact, decomposed skill representation, we  naturally obtain a \textit{partially assistive} robot teaching curriculum to facilitate learning: the human student learns the sub-skills one at a time, and the robot teacher assists with the sub-skills not yet learned, to complete the target task. 
While the robot  assists the human in the teaching task, its behavior differs from those in  common collaborative human-robot interaction tasks~\cite{Dragan2013,ReddyDL18}.
There, the primary objective is to complete the task, and  the robot is fully assistive: if the human does not perform, the robot then tries to complete the task on its own, if possible. In the teaching task, the robot is partially assistive and usually avoids assisting with the specific sub-skill to be learned, in order to encourage student exploration and learning.

As a first attempt, 
we conducted human-subject experiments on two challenging human-robot collaboration tasks, \overcooked{} and \ballmaze{} ( Fig.~\ref{fig:exp_setup}). 
Our results show that COACH enables the robot teacher to model and reason over adaptive human students in cooperative teaching. Also,  a fully-assistive teacher may impede student learning, and a partially assistive teacher indeed motivates the student to explore new strategies.

%% file: sections/2_related_work.tex
\section{Related Work}

\begin{figure}
    \centering
    \includegraphics[width=.95\columnwidth]{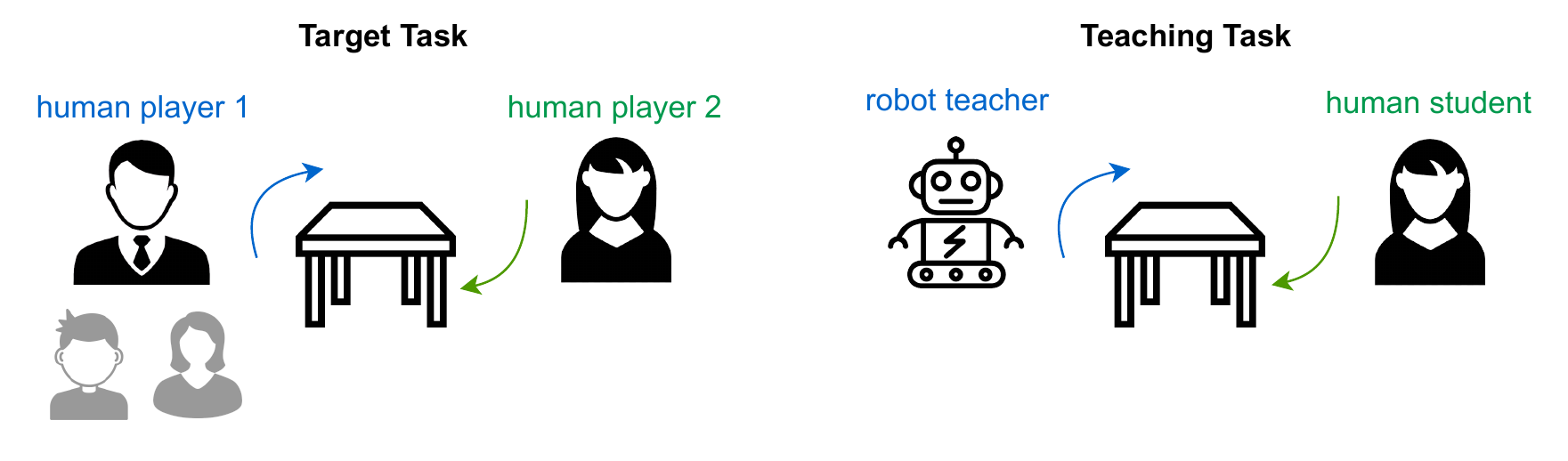}
    \caption{Cooperative robot teaching. In the  target task (left), two human players jointly reorient a table, for example. In the corresponding teaching task (right), the robot teacher interacts with the human student and teaches cooperative skills so that the student learns to cooperate with partners with varying capabilities or preferences in the target task.
    }
    \label{fig:overview}
\end{figure}
\noindent\textbf{Assistance in HRI.} One major aspect of HRI is how the robot could assist humans with a hidden human objective~\cite{Dragan2013,ReddyDL18}. 
The objective of the robot is to infer the human's intention and learns to assist the human. In its simplest form, the action selection and human intention inference are separated~\cite{Leike2008,jeon2020,chris2017}. A decision-theoretic framework, assistant POMDP, is developed to capture the general notion of assistance in HRI~\cite{Fern2014}. The robot integrates the reward learning and control modules to perform sophisticated reasoning over human feedback~\cite{shah2021benefits,sidekick_2021}. However, these two approaches neglect human learning/adaptation and may hinder humans from improving their skills. Our work focuses on how to generate behaviors that facilitate human learning during interactions. 

\noindent\textbf{Collaboration in HRI.} Another important aspect of HRI is to model interactions as the collaboration between the human and the robot~\cite{collaborativeplan}, for which the human and the robot share the same objective.
However, the joint optimal policy, e.g. rotating the table counter-clockwise, is unknown to both agents in the first place. Their interaction is mutually adaptative~\cite{Nikolaidis2016, Nikolaidis2017, Nikolaidis2017hri}. Particularly, as pointed out in~\cite{cirl}, if one side is only aware of partial information about the task, the optimal policy pair naturally induces the behavior of active teaching, active learning, and efficient communication between the robot and human. In this work, we focus on the following setting: given that the robot teacher knows the policy to teach, and how to carry out active teaching.

\noindent\textbf{Teaching Algorithm for Algorithms.} Teaching for algorithms aims to facilitate the learning of the algorithm by choosing or generating training samples. Various teaching techniques including curriculum learning~\cite{cirriculumlearning} and machine teaching~\cite{mt,mt1,mt2,mt3,imt} have been effectively applied to supervised learning and semi-supervised learning problems. Similar ideas are further extended to train reinforcement learning agents to learn complex skills, \eg{}, generate training environment for reinforcement learning~\cite{Gur2022,Portelas2019,Fontaine*2021}, choose various demonstrations~\cite{Brown2019} or learn to decompose the skill~\cite{brian2004, GonzlezBrenes2013WhatAW}. Teaching in cooperative multi-agent RL allows agents to simultaneously become teachers and students for each other~\cite{Omidshafiei2019, Kim2020, Jaques2019}. However, such approaches generally require relatively more data for training and to some extent the controlled learning behavior of the learner. Transfer of these approaches to human learning is promising but difficult.

\noindent\textbf{Teaching Algorithm for Human.} Despite the aforementioned
practical challenges, some algorithms have been successfully deployed for
human learning. Attempts on teaching the crowd on classification or concepts
prove to be
successful~\cite{pomdpteaching,Singla2014,Zilles2011,Doliwa2010, Aodha_2018_CVPR}. While humans
can learn concepts from visual or verbal examples, complex skills like motor
control skills can hardly be mastered through these signals. Recently, skill discovery techniques in reinforcement learning have been introduced to generate a curriculum based on skill decomposition and facilitate humans to learn motor control skills~\cite{srivastava2022assistive}. It focuses on how to adaptively decompose the skill into learnable sub-skills for a human to practice on its own and achieves promising results. Here, we seek to automate the teaching process for humans to cooperate in a physical task and provide a framework for this teaching mode, \eg{}, table co-reorientation.

%% file: sections/3_formulation.tex
\section{Cooperative Robot Teaching}\label{sec:form}

We  identify four key elements in COACH: (1) target task, (2)  student, (3)  teacher, and (4) interactive teaching-learning. 

\noindent\textbf{Target task.} 
In this work, we focus on teaching in \textit{a two-player cooperative task}, which we call it the \textit{target task}.

\begin{defn}\label{defn1}. The \textit{target task} is a two-player cooperative Markov game $\mathcal{M}=( \statespace, \Aactionspace, \Bactionspace, \Tdots, \Rdots, \gamma )$ between two agents, \agenta{} and \agentb{}, where 
\begin{tightlist}
 \item \statespace\  is a set of target task states;
 \item \Aactionspace\ is a set of actions for  agent \agenta;
 \item \Bactionspace\ is a set of actions for  agent \agentb;
 \item $\Tdots(\state' | \state, \Aaction, \Baction)$ is a conditional probability  function on the next target task state $\state'\in \statespace$, given the current state $\state\in \statespace$ and   both agents' actions $\Aaction\in\Aactionspace$ and $\Baction\in \Bactionspace$;
 \item \Rdots(\state, \Aaction,\Baction) is a target task reward function that maps the target task state and players' actions to a real number;
 \item $\gamma$ is a discount factor. 
\end{tightlist}
\end{defn}
At each step $t$, agent \agenta{} and \agentb{} both observe the current task state $\state_t$ and  select their respective actions $\Aaction_t \sim \Apolicy{}$ and $\Baction_t \sim \Bpolicy{}$, where $\pi^i$ the policy of agent $i \text{ for } i=1,2$. They then receive a joint reward $r_t=R(\state_t,\Aaction_t, \Baction_t)$. The next state is updated as $\state_{t+1} \sim \Tdots (\state_{t+1}~|~\state_t,\Aaction_t,\Baction_t)$.

Given the definition of the target task, we first answer how to represent the knowledge/skills. In this work, we choose to represent a \textit{skill} by a policy \optimalpolicy{} to the target task. 
For example, in the table co-reorientation task, the agent needs to learn to deal with either stubborn or adaptive partners. We recognize that there are other ways to represent knowledge/skills, such as a set of demonstrations and the ground-truth reward function. However, such representations are indirectly linked with the skill's performance; therefore, evaluating its proficiency is more obscured. We choose a known policy to be taught as the representation since it can be directly optimized over and evaluated.

\noindent\textbf{Student.} The student policy is non-stationary since it will improve along with teaching. 
We model this evolutionary behavior with a tuple of student policy \Spolicy{} and an
updating function \Supdate{}, $( \Spolicy{}, \Supdate{} )$. The student policy represents the student knowledge state. It will take in the current target task state $s$ as input and output the student's action. The updating function $\Supdate{}$ models how the student changes its policy after each teaching step.

\noindent\textbf{Teacher.} We define the teacher as a knowledgeable agent (expert) who 
knows a set of policies $\Phi^* $
for a target task. The teacher aims to acquire a teaching policy \Tpolicy{}  that can teach any $\optimalpolicy_i \in \Phi^*$ to the student effectively. In this general setting, the choice of the student policy to teach $\optimalpolicy$ depends on the capability, preference, and current knowledge level of the student. 
A principled approach to selecting the policy to teach needs to consider the student's preference, his/her update model for the knowledge level, and an estimate of his/her current capability. In this paper, we assume that we have an oracle to choose the policy to teach $\optimalpolicy{} \in \Phi^*$, such that this policy $\optimalpolicy{}$ matches the preference of the student.
The teacher can be described by a tuple of a target task policy and the corresponding teaching policy, $(\optimalpolicy{}, \Tpolicy{})$.

\noindent\textbf{Interactive teaching-learning}.
In the target task, the teacher knows the target task policy $\optimalpolicy{}$ while the student does not. The teacher's goal is to act in the most informative way so that the student learns $\optimalpolicy{}$ fastest. The choice of $\optimalpolicy{}$ should account for the student's preferences. To embed the objective of teaching and distinguish it from the \emph{Target Task}, we define it as the \emph{Teaching Task} in the following way:
\begin{defn}\label{defn2}. 
Given a target task $\mathcal{M}=( \statespace, \Aactionspace, \Bactionspace, \Tdots, \Rdots, \gamma )$, a  student $( \Spolicy{} , \Supdate{} )$, and a policy to teach $\optimalpolicy{}$ for the target task, 
the teaching task is a POMDP   $\bar{\mathcal{M}} = ( \PomdpStateSpace, \PomdpActionSpace, \PomdpTransition, \PomdpObservation, \PomdpObservationFunction, \PomdpRewardFunction, \PomdpGamma )$ for the teacher, where
\begin{tightlist}
 \item \PomdpStateSpace\ is a set of teaching  states:  \PomdpState{} ~= $(s, \Spolicy)$, for target task state $s\in\statespace\ $ and student policy $\Spolicy$; 
 \item \PomdpActionSpace\ is a set of actions: $\PomdpActionSpace = \Aactionspace \cup \Bactionspace$;
 \item $\PomdpTransition(~\PomdpState' |~\PomdpState, \PomdpAction)$ is a conditional probability  function on the next state $\PomdpState' \in \PomdpStateSpace$, given the current state $\PomdpState\in\PomdpStateSpace$ and teacher's action $\PomdpAction\in\PomdpActionSpace$;
 \item \PomdpObservation\ is a set of observations:  $\SmallPomdpObservation{} = ( s, r )$, for target task state $s\in\statespace$ and target task reward $r$;
 \item $\PomdpObservationFunction{}(~\SmallPomdpObservation{}~ |~\PomdpAction{}, \PomdpState{})$ is a conditional probability function on the observation  $\SmallPomdpObservation{}\in\PomdpObservation\ $, given teacher's action $\PomdpAction\in\PomdpActionSpace$ and current state $\PomdpState\in\PomdpStateSpace$;
 \item $\PomdpRewardFunctionDots(\PomdpState{}, \PomdpAction{} , \PomdpState{}')$ is a teaching  reward function that maps  current state $\PomdpState{}\in \PomdpStateSpace{}$, teacher's action $\PomdpAction\in\PomdpActionSpace$, and next state $\PomdpState'\in \PomdpStateSpace{}$ to a real number measuring the effectiveness of teaching; 
 \item \PomdpGamma\ is a discount factor.
\end{tightlist}
\end{defn}

The objective of the teaching task is to derive a teaching policy \Tpolicy{}, enabling students to learn $\optimalpolicy{}$ for the target task fastest.
More specifically, the teacher can influence the student through interactive actions $\PomdpAction \in \PomdpActionSpace$. 

First, we define the learning behavior of the student. We consider humans would take the interaction history into account. The history of observation is $h_t = [(\state_0,r_0),...,(\state_t,r_t)]$. Thus, the student updates \Spolicy{} with any arbitrary iterative functions conditioned on the history of interactions:
$\Spolicy_{t+1}=\Supdate(\Spolicy_t,h_t)$.

Next, we give the definition of the reward function. To incentivize the teacher to speed up the teaching process, we introduce a step-wise teaching cost to the teacher $c_t= \Cdots(s_t, \PomdpAction_t)$ to penalize unnecessary teaching actions. To this end, we define the reward function as
\begin{equation}
\small
\PomdpRewardFunction(\PomdpState, \PomdpAction_t, \PomdpState'; \PomdpDistance,\Cdots,  \optimalpolicy{}, \omega) = \PomdpDistance(\Spolicy_t, \optimalpolicy{}) - \PomdpDistance(\Spolicy_{t+1}, \optimalpolicy{}) - \omega \Cdots(s_t, \PomdpAction_t),
\label{pomdpreward}
\end{equation}where  $\PomdpState = (\state_t, \Spolicy_t),~ \PomdpState' = (\state_{t+1}, \Spolicy_{t+1})$, $\omega$ is the weighting factor to trade-off the teaching cost and teaching efficiency, and {$\PomdpDistanceMeasure$ can be any reasonable distance measure between two policies, \eg{}, initial state value in the target task.}

Lastly, we introduce our choice of the teaching policy $\Tpolicy{}$. To devise a student-aware teaching strategy, apart from the current state $\state_t$ and the target policy \optimalpolicy{}, our $\Tpolicy$ also takes the history of observation as input. The action of the teacher can be sampled from the policy,
i.e., $\PomdpAction_t \sim \Tpolicy(\PomdpAction_t ~| ~ h_{t-1},s_t, \optimalpolicy{})$.  The solution to the POMDP $\mathcal{M'}$ is a teaching policy $\Tpolicy{}$ that maximizes the expected sum of rewards  $\mathop{\mathbb{E}}_{\PomdpAction_t \sim \Tpolicy}[\sum_{t=0}^{\infty}\PomdpGamma^t \PomdpRewardFunction(\PomdpState, \PomdpAction_t, \PomdpState')]$.

%% file: sections/4_method.tex
\section{Method}
\begin{wrapfigure}{r}{0.5\textwidth}
\vspace{-25pt}
\hspace{-10pt}
\begin{minipage}{0.5\textwidth}
\begin{algorithm}[H]
\caption{\small Approximated Solution to the Teaching Task}\label{alg:cap}
{\fontsize{9}{9}\selectfont
\begin{algorithmic}[1]
\Require Maximum Interactions $L$, Predefined Interactions $N$
\For {$k \in \PomdpActionSpace$}:
\State Randomly initialize $\lambda$ and $\alpha_t$, $\beta$, and $\dataset = \{\}$
\For {\texttt{$i= 1,2,...,N$}}:
\State $\dataset.\texttt{add}(v_i)$
\EndFor
\EndFor
\For {\texttt{$i= 1,2,...,L$}}:
\For {$k \in \PomdpActionSpace$}:
\State Learn $\lambda$ and $\alpha_t$, $\beta$ from $\dataset$
\EndFor
\State $k \gets$  Action selection from $\lambda$ and $\alpha_t$, $\beta$ 
\State $v_i \gets $ Performance measure from interactions
\State $\dataset.\texttt{add}(v_i)$
\EndFor
\end{algorithmic}}
\end{algorithm}
\vspace{-20pt}
\end{minipage}
\end{wrapfigure}
In this section, we provide a solution that grounds all the elements in the conceptual framework of COACH. The main spirit of our solution is to parameterize students' knowledge state with IRT and decompose complex tasks into a set of role-based independent skills.
The action space in definition~\ref{defn2} allows the teacher to take all possible actions in the 2-player task. Thus, the teacher is able to switch roles freely. For example, the teacher may serve as either follower or leader in the classic leader-follower model~\cite{kheddar2009,kheddar2011}. 
This enables easier evaluation of the  students' proficiencies and provides a ground to derive the partially assistive interaction mode.
Our solution is summarized in Algorithm~\ref{alg:cap}.
To begin with, we first define the action space, $\PomdpActionSpace$.

\subsection{Action}\label{sec:action_space}

The action space is constructed through sub-skill decomposition. 
Sub-skills decomposition is well-studied for single-agent tasks~\cite{Graves2012label,shiarlis2018,kipf2019compositional}. However, extending the same idea to a multi-agent setting is still challenging since task completion relies on the interaction among multiple parties.
We observe that in a multi-agent game, the task naturally comprises several roles, of which each agent takes a subset.
The well-established leader-follower model is a particular choice of role-based skill decomposition~\cite{kheddar2009,kheddar2011, natha2012framework,roleofrole}. 
Therefore in our work, we propose to decompose skills based on role allocation.
We divide the skill into $K$ independent teachable sub-skills according to the student's potential roles in the task. 
The teacher's action space $\PomdpActionSpace = \{ k: k\in \mathbb{Z},~ 0 \leq k < K \}$ consists of teaching each sub-skill. Such a decomposition of skills naturally leads to a partially assistive mode of interaction. 
\subsection{State} \label{state_space}
The state space is constructed with Item Response Theory (IRT)~\cite{irt}. IRT provides a parametric form to represent students' skill levels. Given the limited interactions, we adopted the simplest form, the one-parameter logistic model (1PL), to model human skills. In the 1PL model, each sub-skill $k \in \PomdpActionSpace$ is assigned a parameter $\beta^k$ representing the difficulty, and a parameter $\alpha^k$ called the \emph{proficiency} representing a student’s knowledge state. 
The probability that a student has mastered sub-skill $k$ is given by $P(k) := \sigma(\alpha^k - \beta^k)$, where $\sigma$ is the sigmoid function. Hence, instead of representing the state with the student's policy $\Spolicy{}$, we use $(\alpha,\beta)^K$ to represent the hidden state. That is, for $\PomdpState \in \PomdpStateSpace, \PomdpState = (\state, (\alpha, \beta)^K)$, where $(\alpha ,\beta)^K$ is hidden. For each student and each $k\in \PomdpActionSpace$, we assume that $\alpha$ changes over time while $\beta$ does not. 
\subsection{Transition}

The transition model consists of two main parts, the target task transition model $\Tdots$, and the student's update function $\Supdate$. While the former one is known to the teacher, we need to make assumptions about the latter one. Since we define the state space over the student's proficiency $\alpha$ in Sec~\ref{state_space}, the transition model is also constructed over the proficiency.
Following the previous work on online estimation of student proficiency ~\cite{Ekanadham2017TSKIRTOE, Wilson2016BackTT}, for each sub-skill, we model the student's proficiencies over time as a Wiener process:
$
\small
    \Supdate(\alpha_{t+\TimeInterval} | \alpha_t) = \exp\left(
        -\frac{(\alpha_{t+\TimeInterval} - \alpha_t)^2}{2\lambda \TimeInterval}
    \right),
    \label{eq:subskill_prof_update}
$
where $\TimeInterval{}$ refers to the step interval and $\lambda$ is a parameter controlling the “smoothness” with which student's proficiency varies over time. For each student and for each $k\in \PomdpActionSpace$, we assume $\lambda$ does not change over time and is learned for each sub-skill respectively. To this end, we construct the transition model in the POMDP as $\PomdpTransition = \{\Tdots, U\}$, where $\Tdots$ is the transition function in the target task. 

\subsection{Observation}

The observation is composed of the target task state and the reward received, $(s, r)$. Recall that in Sec~\ref{sec:action_space}, we define the action as choosing one sub-skill to train the student, which is a macro-action. 
For teaching sub-skill $k$, we
redefine the observation as the ratio between \ttt{} rewards achieved by the student's current and the policy to be taught:
$\small
    v
    := \frac{
        R(\state, \PomdpAction, \Saction)
    }{ 
        R(\state, \PomdpAction, a^*)
    },
    \label{eq:subskill_prof}
$
where $a^*$ is the action generated by the policy 
 to be taught $\optimalpolicy$ and $a^\mathrm{S}$ is the action from student's policy given the same \ttt{} state $s$. Since all the sub-skills are treated equally, we will omit the index $k$ for simplicity in the following discussion. 
As a result, for $\SmallPomdpObservation \in \PomdpObservation$, $\SmallPomdpObservation = (s, v)$. Unlike the binary response in conventional knowledge tracing, the response $v$ we have is continuous and we assume the teacher will only teach one sub-skill at a time. Thus, we use the continuous Bernoulli distribution to construct the observation model:
$\small
    Z(v | P(k)) = P(k)^{v}(1-P(k))^{1-v},
$
where $k$ is the sub-skill being taught when $v$ is observed. 
As a result, the observation model can be defined as $\PomdpObservationFunction = \{I, Z\}$, where $I$ is an identity mapping for the observable target task state, $I(s) = s$.

\subsection{Reward}
The distance between the student's policy and the policy to be taught can be represented using $P(k)$. We represent the distance as the average of one minus master probabilities of each sub-skill:
$\small \PomdpDistance (\Spolicy, \optimalpolicy) = \frac{\sum_{k=0}^{K}1- P(k)}{K}$.
 There are other ways to specify the goal according to the decomposition of the skill, \eg{} weakest or multiply~\cite{pittir26017}. We choose the sum due to our independence assumption on sub-skills. In this work, we assume the cost is uniform, thus, given a finite horizon of interactions, maximizing the reward function defined in Equation~(\ref{pomdpreward}) is equivalent to maximizing $\small \PomdpRewardFunction(\PomdpState, \PomdpAction_t, \PomdpState') = \frac{\sum_{k=0}^{K} P_{t+1}(k) - P_t(k)}{K}$, where $P_{t}(k) = \sigma(\alpha^k_t - \beta^k)$.
\subsection{Model Learning and Decision Making}
We use the student's performance during the interactions to estimate both $\lambda$ and $\alpha_t, \beta$. Parameters for each sub-skill are learned separately, thus, we omit $k$ for simplicity. Let $v_{1:t}$ denote sequences of student's performance measure against the policy to be taught up to step $t$. We have the posterior $P(\lambda, \alpha_t, \beta |v_{1:t}) \propto P(v_{1:t}|\lambda, \alpha_t, \beta) P(\lambda, \alpha_t, \beta)$.
The conditional probability of the observation and current proficiency can be obtained by integrating out all the previous proficiencies. The likelihood can be approximated through
     $P(v_{1:t}|\lambda, \alpha_t, \beta) \approx 
     \prod_{t'=1}^t \int P(v_{t'}| ~\lambda, \alpha_{t'}, \beta)\Supdate(\alpha_{t'}|\alpha_t) \mathrm{d}\alpha_{t'}.$
An approximation of the log posterior over the student’s current proficiency given previous responses can be derived to learn the parameters $\lambda$ and $\alpha_t$, $\beta$. 
Following~\cite{Ekanadham2017TSKIRTOE, Wilson2016BackTT}, we employ maximum a posteriori estimation (MAP) to learn these parameters.
Given the estimation of the current state using the past history, we use a one-step look-ahead. Such a choice allows us to reduce the impact of the learned inaccurate model and generate a more efficient solution compared with the full-blown POMDP solution.
At timestep $t$, the teacher's action is given as
\begin{equation}
    \PomdpAction_{t+1} = \argmax_{k\in\PomdpActionSpace} \int \Supdate(\alpha^k_{t+1} | \alpha^k_t) 
    P_{t+1}(k)~\mathrm{d}\alpha^k_{t+1} - P_{t}(k).
\end{equation}
In practice, the student is asked to perform on each sub-skill for a few interactions to initialize the parameters. 
\subsection{Training on Sub-skills}

Our overall strategy for training students on each sub-skill is to diversify scenarios the student would encounter during training. Training students on sub-skills naturally leads to a partially assistive partner on unlearned sub-skills, which allows the student to explore the sub-skill freely. 
We adopt an intuitive assumption: \emph{an agent learns cooperation better with a diverse group of partners}. Such a teaching strategy is effective when dealing with synthetic students~\cite{lupu21a, Zhao2021}. The student could learn from a diverse set of partially assistive partners or learn to cope with them by acquiring new skills.

%% file: sections/5_experiment.tex
\section{Experiments}
 We carried out two human-subject experiments to demonstrate how COACH works, one in simulation (Overcooked-AI~\cite{Carroll2019OnTU}) and the other with a real robot (\ballmaze{}). Experiment setups are shown in Figure~\ref{fig:exp_setup}. We investigated the teaching performances of three types of teachers: 
the \textbf{fully-assistive} teacher who performs optimally concerning the student's initial capability, the \textbf{student-aware} teacher who behaves according to our teaching strategy, and the \textbf{random} teacher. The random teacher in the \ballmaze{} experiment chooses sub-skills randomly, while the random teacher in the \overcooked{} experiment executes actions randomly.
\subsection{Setups}
\noindent\textbf{Overcooked-AI.} Overcooked-AI is a benchmark environment for fully cooperative human-AI task performance and has become a well-established domain for studying coordination~\cite{knott2021eval,Charakorn2020InvestigatingPD, nal2021intention, sarkar2021pantheonRL}. The goal of the game is to cook and deliver as much soup as possible in a limited time. 
We decompose the policy into two sub-skills: \emph{putting ingredients in the pot} and \emph{delivering the soup}. To put ingredients in the pot, there exists one \emph{efficient strategy }to pass the ingredient through the middle table. 
In brief, rather than picking up one onion at a time and putting them into the pot, the efficient strategy is 1) put multiple onions on the middle table; 2) go to the pot; 3) pick up onions from the middle table; 4) put them into the pot. The overall idea is to reduce the number of movements needed to deliver the same amount of ingredients. 
\begin{wraptable}{r}{0.48\columnwidth}
    \centering
    \begin{tabular}{cc}
        \hspace{-10pt}\cincludegraphics[width=0.24\textwidth]{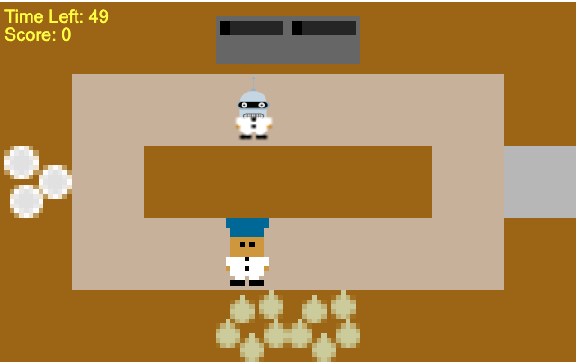}& 
        \cincludegraphics[width=0.2\textwidth]{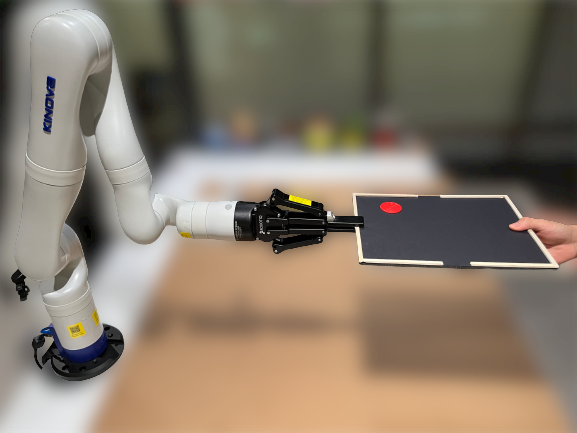}
        \\
        \hspace{-10pt}(a) & (b)
    \vspace{-2mm}
    \end{tabular}
    \captionof{figure}{Experiment setups. (a) Overcooked-AI layout: human participants control the ``chef'' and the robot controls the ``robot''. (b) The real robot setup of \ballmaze{} with a simplified setting.}
    \label{fig:exp_setup}
\end{wraptable} 
We recruited $N$=20 (8 females and 12 males) participants and randomly assigned them into three groups, each with a different teaching strategy. Students are trained with different teachers and are evaluated with a common unseen partner. We emulate the human partner in evaluation using a trained model. 
Each participant was trained for 5 games and then evaluated for 1 game. 

\noindent\textbf{Cooperative Ball Maze.} The \ballmaze{} game requires coordination from both the robot and the human. Each party will hold one side of the maze board and tilt it to move the ball out from one of the two exits. 
We define two sub-skills \emph{leading the rotation} and \emph{following the rotation}.
We recruited $N$=21~(10 females and 11 males) participants to carry out human-subject experiments.
The participants were first evaluated in the two sub-skills, then trained for 20 interactions, and finally re-evaluated in the two sub-skills. Details can be found in the supplementary materials.
\begin{table}[t]
    \centering
    \begin{tabular}{ccc}
        \cincludegraphics[scale=0.2]{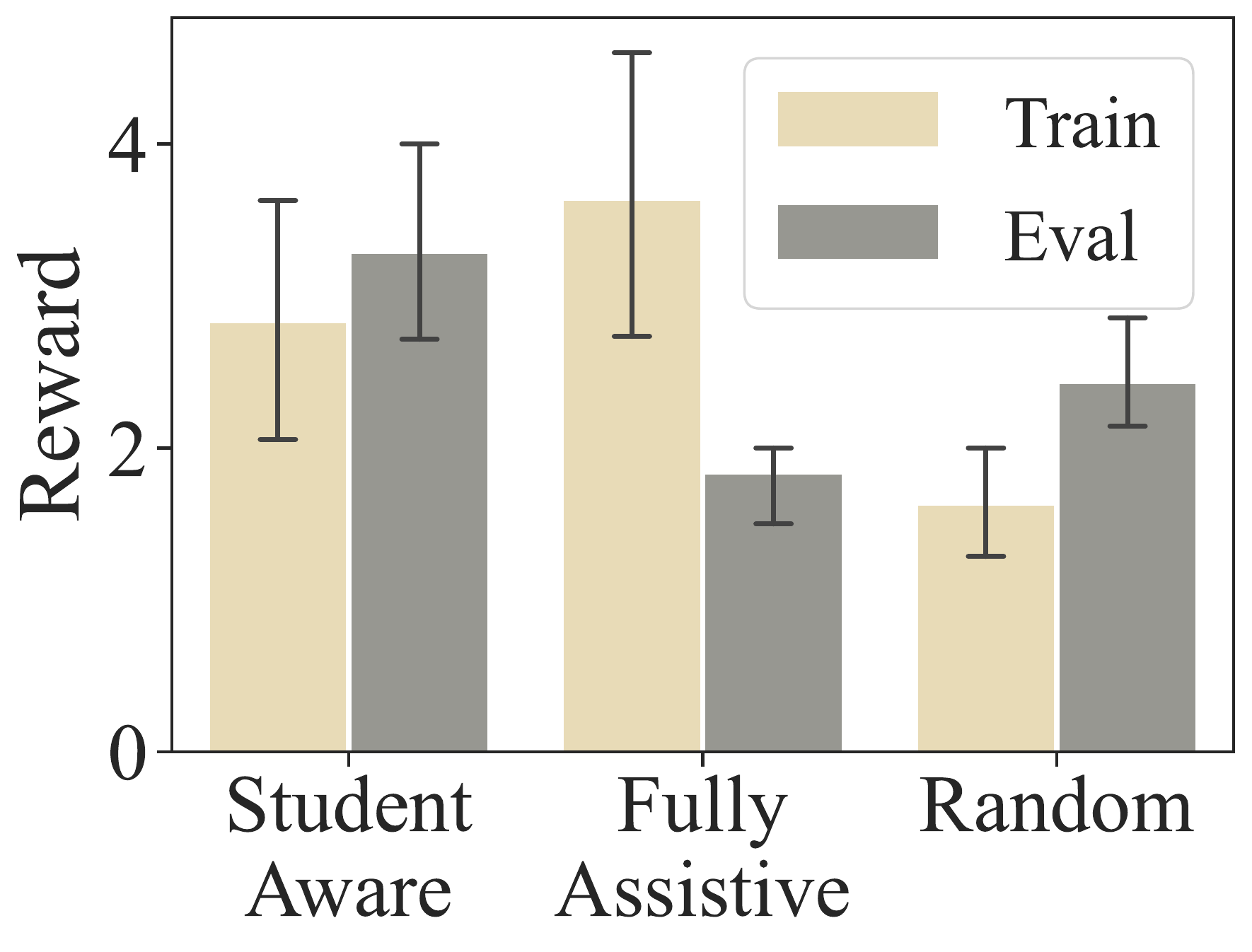}& 
        \cincludegraphics[scale=0.2]{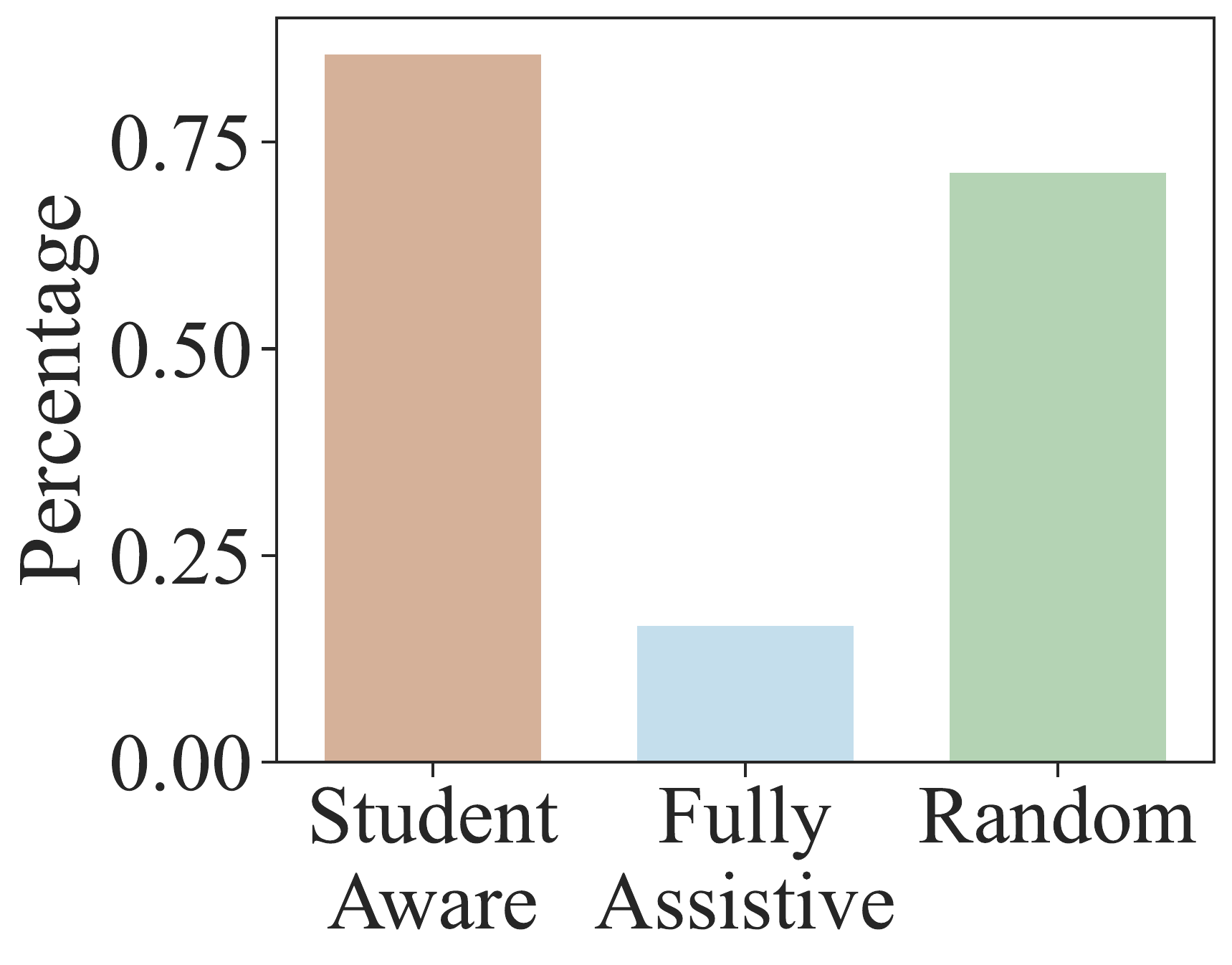}&
        \cincludegraphics[scale=0.2]{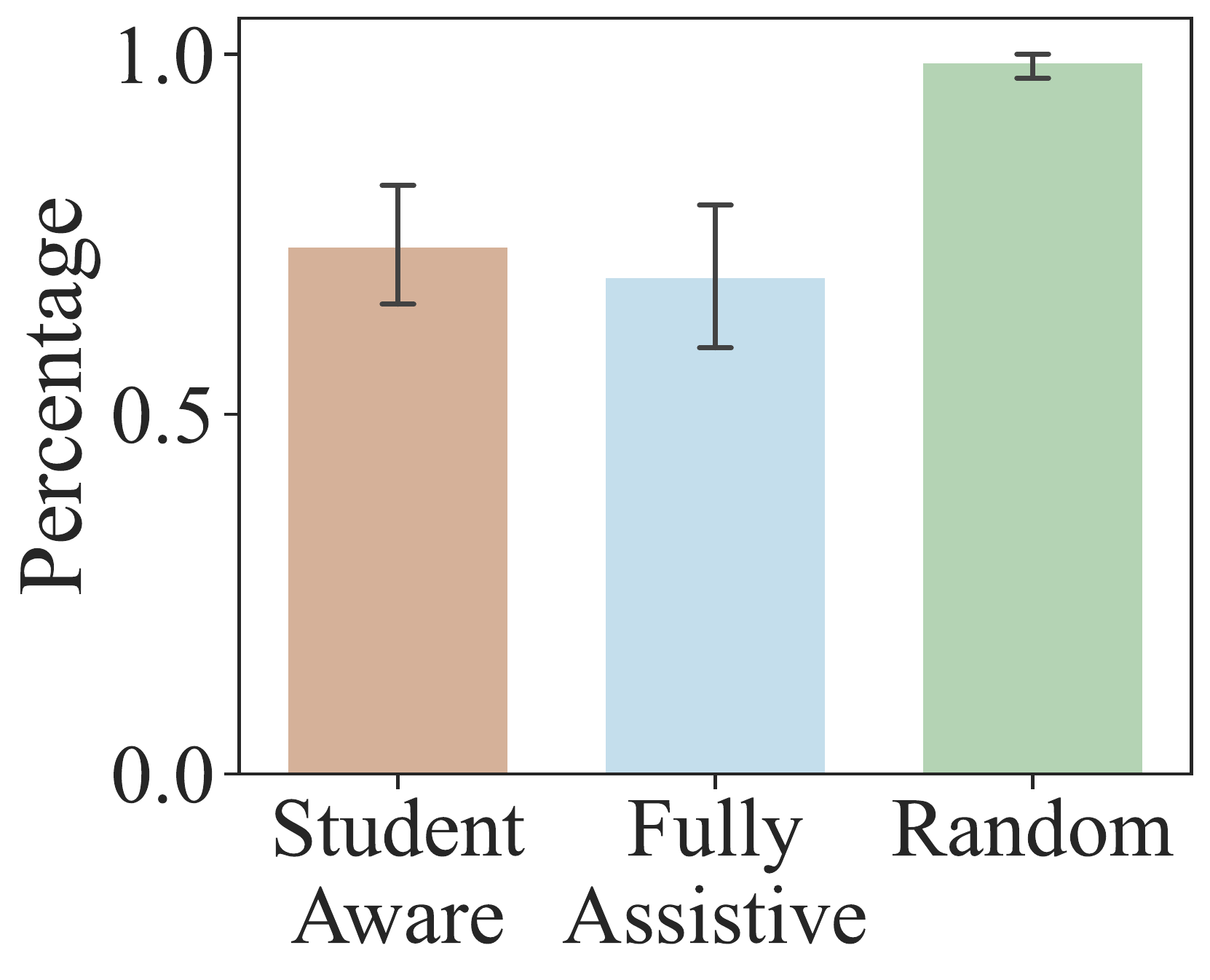}
        \\
        \hspace{12pt}(a) & \hspace{22pt}(b) & \hspace{22pt}(c)
    \end{tabular}
    \captionof{figure}{Results of the Overcooked-AI experiment.
    (a) Rewards achieved together by the human-robot pairs during training and evaluation. The error bars correspond to the 95\% confidence intervals (95\%CI). The student-aware teacher outperformed the fully assistive and the random teachers in terms of the evaluation reward (with one-sided $p$-values $0.001$ and $0.03$).
    (b) Percentage of students who found the efficient strategy. None of the students are aware of this strategy at the beginning of the training.
    (c) Percentage of reward achieved by the human participants during training.
    }
    \label{fig:overcooked_result}
\end{table}
\begin{table}[t]
    \centering
    \begin{tabular}{ccc}
        \cincludegraphics[scale=0.2]{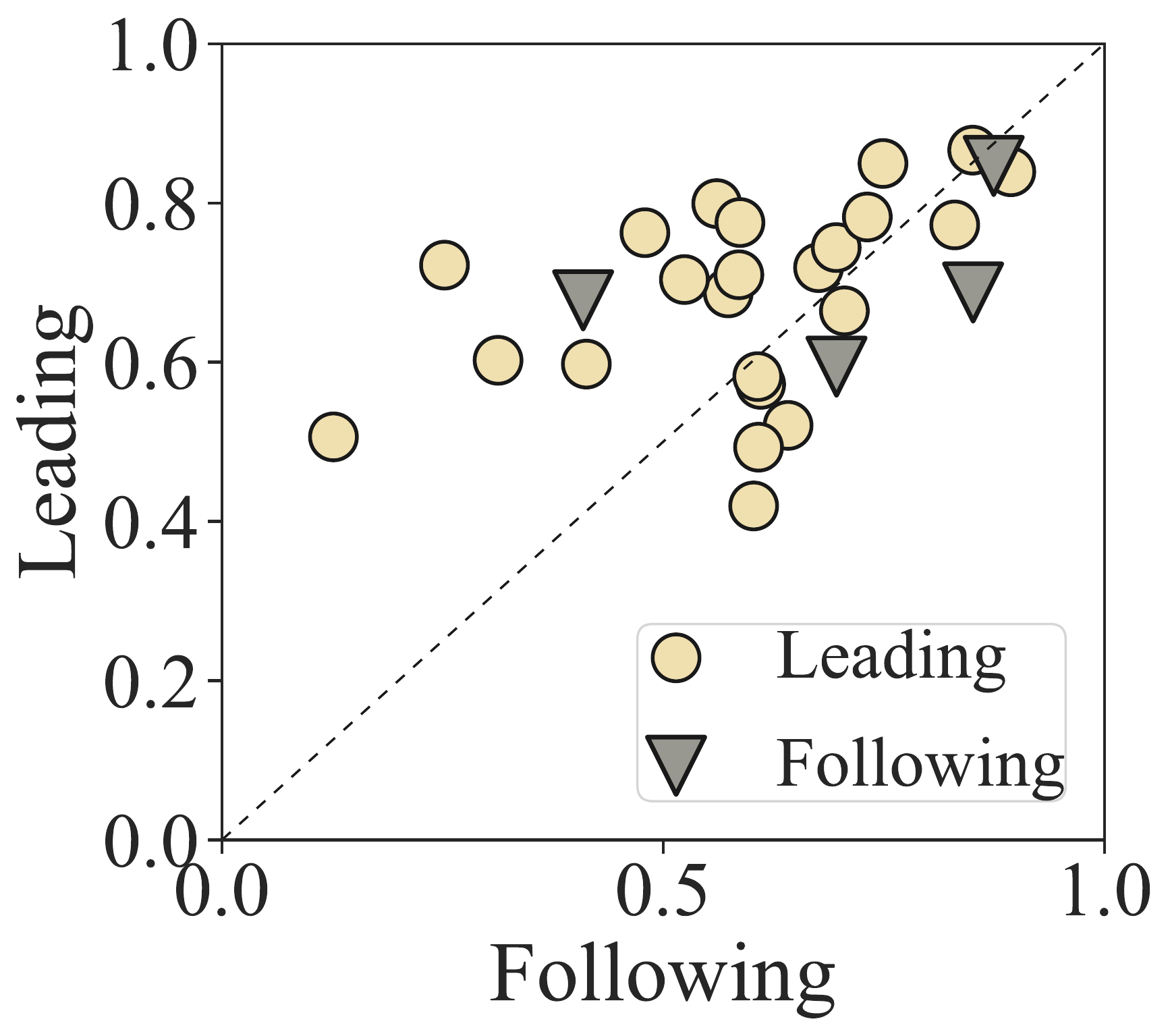}& %
        \cincludegraphics[scale=0.2]{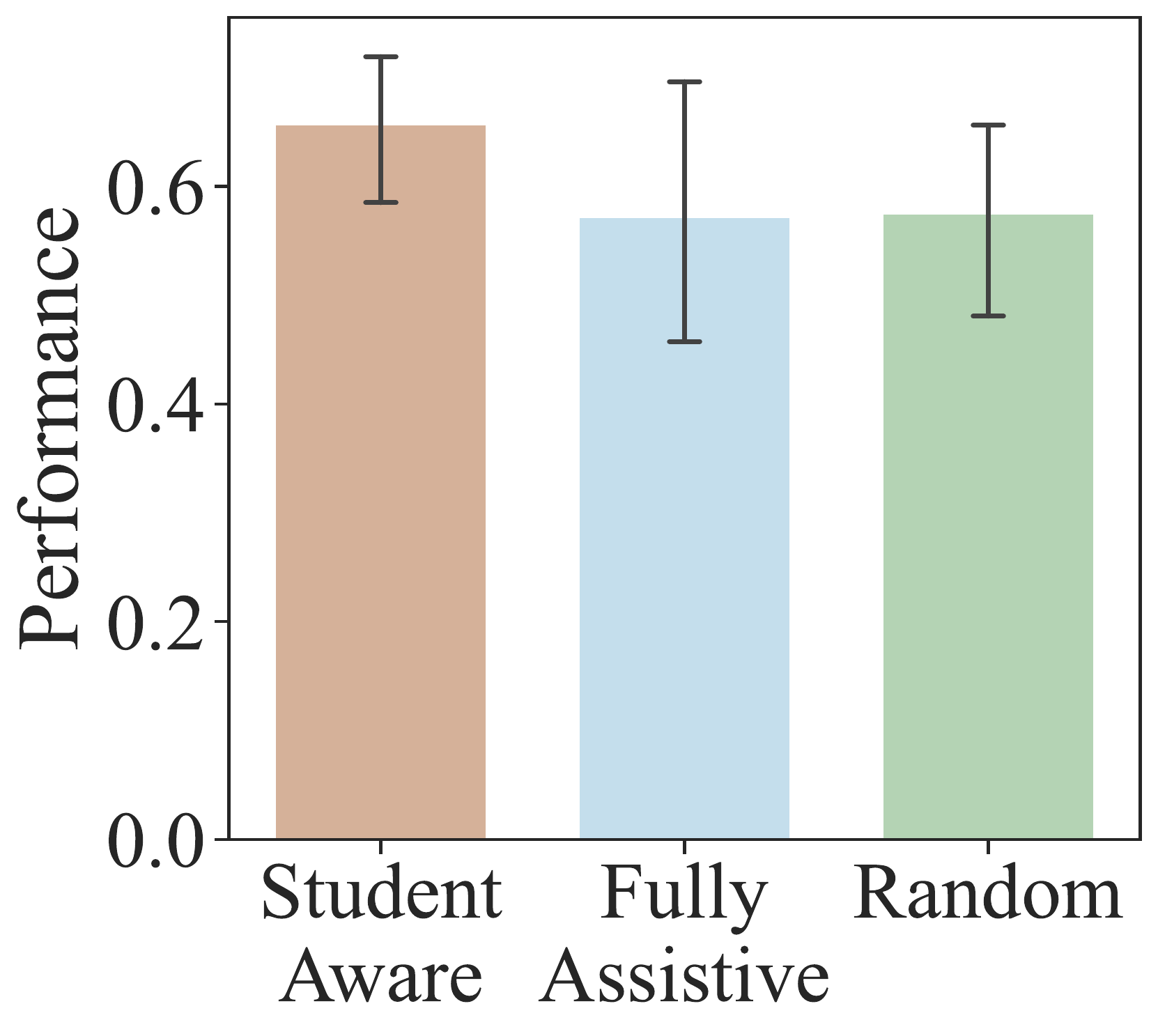}&
        \cincludegraphics[scale=0.2]{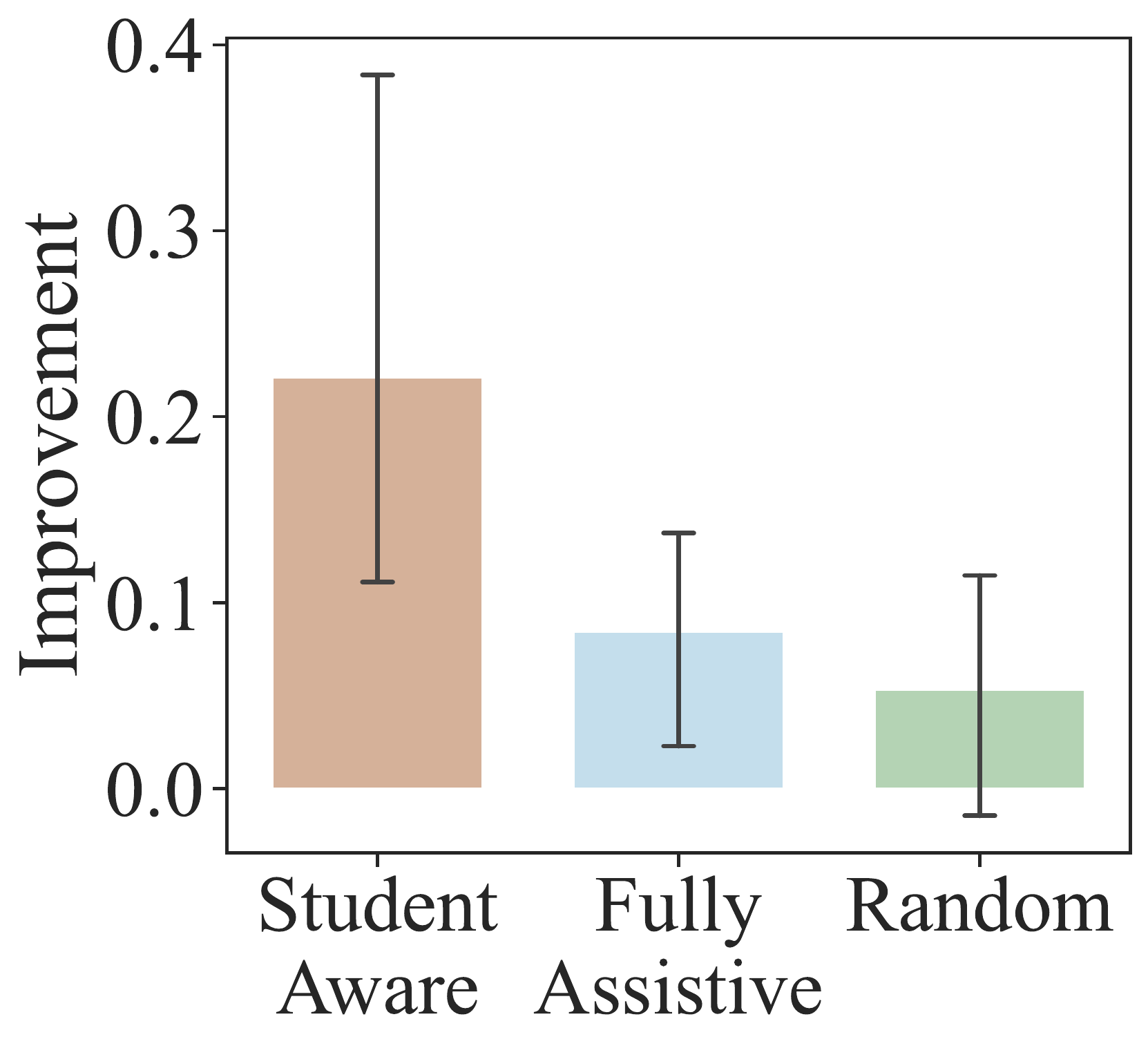}
        \\
        \hspace{12pt}(a) & \hspace{22pt}(b) & \hspace{22pt}(c)
    \end{tabular}
    \captionof{figure}{Results of the Cooperative Ball Maze experiment.
    (a) Evaluation of performances of the two sub-skills of all participants. The marker styles correspond to the sub-skill preferences of the participants.
    (b) Evaluation performances. The error bars correspond to the 95\%CIs.
    (c) Improvements after 20 interactions. The error bars correspond to the 95\%CIs.
    The students improve more under student-aware teachers than both fully-assistive and random teachers (with one-sided $p$-values $0.069$ and $0.039$). 
    }
    \label{fig:ballmaze_result}
\end{table}
\subsection{Results}
\emph{A fully-assistive teacher impedes human's acquisition of skills}. In the Overcooked-AI experiment shown in Figure~\ref{fig:overcooked_result}(a), we observe that the students trained with a fully-assistive teacher perform worse than the students with a random teacher: it seems that a student becomes ``lazy'' and free rides the teacher when the teacher unilaterally adapts to the student and performs optimally. We further investigate the learning pattern of the ``lazy student'' problem and find out that \emph{this ``laziness'' does not lie in the student's reluctance to take actions, but rather in the lack of motivation to explore and improve}. In Figure~\ref{fig:overcooked_result}(c), we show the percentage of reward achieved by the student in Overcooked-AI during training. Compared with the student-aware counterpart, the percentage of reward achieved by humans is similar. However, only 17\% of the participants of the group find out the efficient strategy (Figure~\ref{fig:overcooked_result}(b)), which is crucial to achieving high scores in the evaluation. 

\emph{Partially assistive or random partner motivates students to explore new strategies.} 
By leaving some/all work to the student, partially assistive and random teachers both motivate the student to acquire new skills. This is shown in Figure~\ref{fig:overcooked_result}(b) that most of the students under these two teachers can find out the efficient strategy in Overcooked-AI. However, their performance and the robustness of the learned strategies differ significantly. Though multiple explanations could account for it, we hypothesize the student under the random teacher learns a single fixed strategy to finish the task alone (Figure~\ref{fig:overcooked_result}(c)). Such a strategy that completes the task alone cannot utilize the possibly helpful inputs from the partner, therefore resulting in a poorer performance score.

\emph{An individualized curriculum should be designed for each student.} 
In the post-experiment survey of \ballmaze{}, we asked the participants ``which mode of the robot is easier to cooperate with?''. Out of the 21 participants, 4 participants preferred to follow the robot and 17 participants preferred to lead the robot. Moreover, as we evaluated the student performance with partners of different sub-skills, we found that the student performances were consistent with their declared preferences (Figure~\ref{fig:ballmaze_result}(a)).  That is to say, the student may have a bias over which strategy to acquire, and tailoring the teaching curriculum to focus on that specific strategy is efficient and more intuitive to the student.
\begin{wraptable}{r}{0.4\columnwidth}
    \centering
    \begin{tabular}{cc}
        \cincludegraphics[scale=0.2]{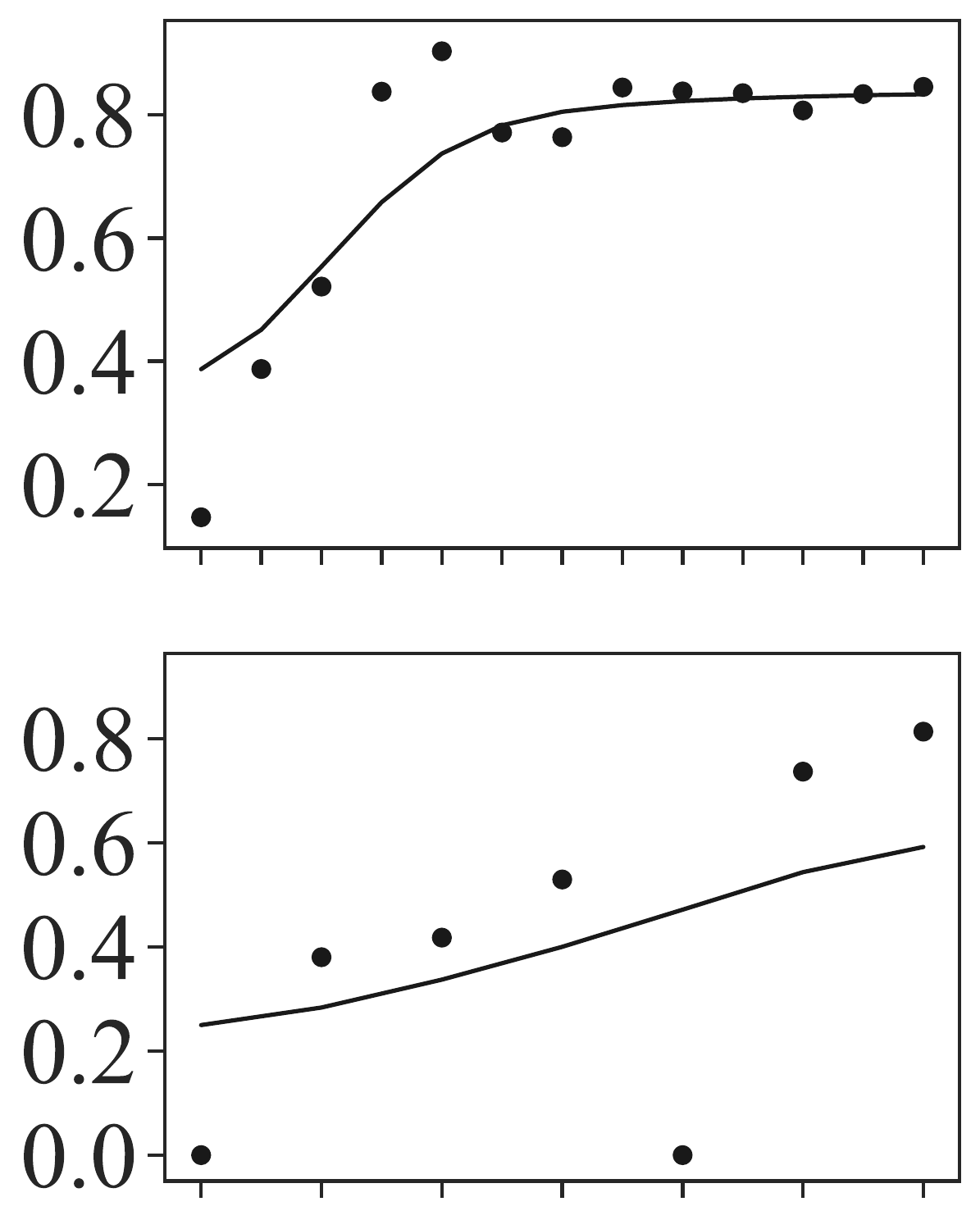}& 
        \cincludegraphics[scale=0.2]{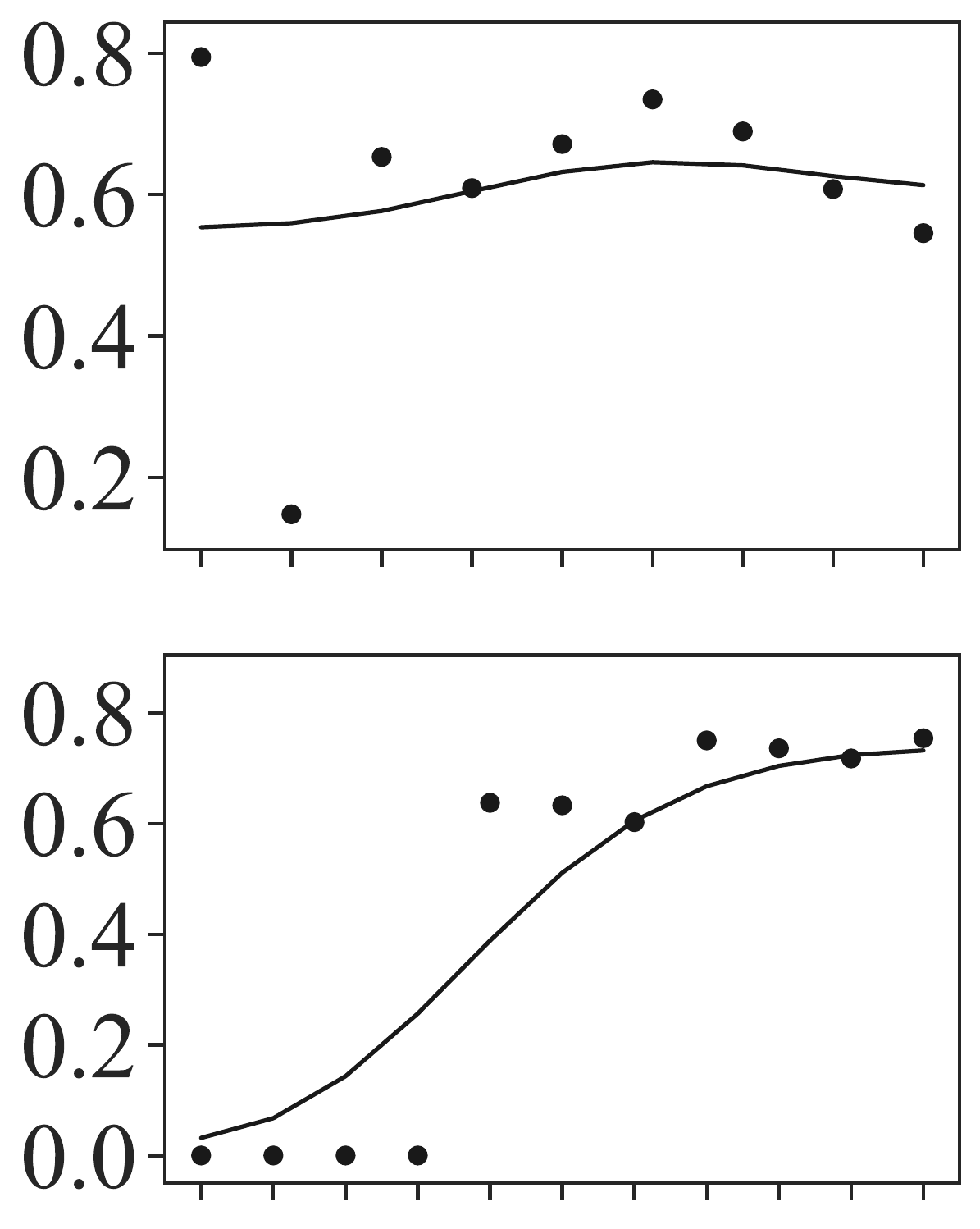}
        \\
        (a) & (b)
    \end{tabular}
    \captionof{figure}{
    Sub-skill performances (vertical axis) with respect to training progress (horizontal axis) of two example participants trained by the student-aware teacher. 
    Dots represent the raw scores and lines represent the smoothed scores.
    The top and bottom figures correspond to leading and following sub-skills respectively.
    (a) Participant 4. The student improved more when trained in the leading sub-skill.
    (b) Participant 6. The student improved more when trained in the following sub-skill.}
    \label{fig:ballmaze_example_participants}
    \vspace{-40pt}
\end{wraptable}
As demonstrated in Figure~\ref{fig:ballmaze_example_participants}(a), after the first 6 trials that estimated the student's proficiency for each sub-skill, the teacher found out this student improved more as the leader,
therefore, the teacher allocated 10 trials to perfect the \emph{leading} sub-skills and only 4 trials for \emph{following}.
Moreover, one participant in the random teacher group responded ``the robot leading mode is too difficult and I gave up''. This demonstrates the importance of an individualized curriculum: though there are multiple equally optimal strategies, the individual may have strong preferences, and teaching a non-preferable strategy will discourage the student from learning anything at all. 
We refer the readers to the Appendix for the complete data of all participants.

%% file: sections/6_limitation.tex
\section{Limitation}
  
\noindent\textbf{Decomposition into sub-skills.} 
For many tasks, it is not easy to identify distinct roles to fulfill the local-independence criteria of sub-skills. We manually decompose the skill into a few sub-skills according to the role of the student. Often, such a decomposition may not be possible or requires careful design. We can mitigate this problem with recent progress on skill decomposition in single-agent task~\cite{srivastava2022assistive} and role-based task decomposition in multi-agent tasks~\cite{wang2021rode}. However, it still demands much more effort to verify their efficacy with a real human on real-world tasks.

\noindent\textbf{Teacher's Knowledge.} In the definition of the teaching task, we assume the teacher has full knowledge of the policies to be taught. However, it can be hard for the robot to know the oracle human policy beforehand. To make the conceptual framework practical, we need to relax the requirement on the teacher's prior knowledge. In our implementation, we reduce such an assumption by approximating the distance through the difference in performances. There can be cases where the target performance is hard to know or such relaxation results in severe information loss. We need more insights on tasks to make the framework practical. 

\noindent\textbf{Curriculum design.} 
In this work, we only design the curriculum over different sub-skills. However, during our experiment, we observe that humans show various responses to the same sub-skill of different difficulties. 
One specific finding is that people may give up learning when the task becomes too difficult. 
As a result, a finer-grained curriculum on the sub-skill training shall be generated to further facilitate human learning.

%% file: sections/7_conclusion.tex
\section{Conclusion}
\vspace{-2mm}
In this work, we propose a conceptual framework, Cooperative Robot Teaching,
that enables robots to teach humans in cooperative tasks. We show that, by
abstracting a teaching task over the original duo cooperative task, the robot
can learn to act as a specialized teacher to humans. To be more specific, we
model the teaching task as a POMDP with hidden student policy and propose a
partially assistive teaching curriculum to support human learning. We believe
that robot teaching fills in the gap in the bilateral knowledge transfer in
HRI: unlike other HRI tasks where the humans instruct the robots how to
behave, the role is reversed and robots try to instill the knowledge back
into humans. Despite the challenges that lie ahead, we believe that
robot teaching has great potential and is a necessary step forward to bring
robots closer to our daily life.

\textbf{Acknowledgments.}
This research is supported in part by the National Research Foundation, Singapore under its Medium Sized Centre Program, Center for Advanced Robotics Technology Innovation (CARTIN), and  AI Singapore Programme (AISG Award No: AISG2-PhD-2022-01-036[T] and AISG2-PhD-2021-08-014), and by the Science and Engineering Research Council, Agency of Science, Technology and Research, Singapore, under the National Robotics Program (Grant No. 192 25 00054).

%% file: main.bbl
\begin{thebibliography}{63}
\providecommand{\natexlab}[1]{#1}
\providecommand{\url}[1]{\texttt{#1}}
\expandafter\ifx\csname urlstyle\endcsname\relax
  \providecommand{\doi}[1]{doi: #1}\else
  \providecommand{\doi}{doi: \begingroup \urlstyle{rm}\Url}\fi

\bibitem[Gillies(2016)]{Gillies2016CooperativeLR}
R.~M. Gillies.
\newblock Cooperative learning: Review of research and practice.
\newblock \emph{Australian Journal of Teacher Education}, 41:\penalty0 39--54,
  2016.

\bibitem[Ross and Smyth(1995)]{Ross1995DifferentiatingCL}
J.~A. Ross and E.~M. Smyth.
\newblock Differentiating cooperative learning to meet the needs of gifted
  learners: A case for transformational leadership.
\newblock \emph{Journal for the Education of the Gifted}, 19:\penalty0 63--82,
  1995.

\bibitem[Omidshafiei et~al.(2019)Omidshafiei, Kim, Liu, Tesauro, Riemer, Amato,
  Campbell, and How]{Omidshafiei2019}
S.~Omidshafiei, D.~K. Kim, M.~Liu, G.~Tesauro, M.~Riemer, C.~Amato,
  M.~Campbell, and J.~P. How.
\newblock {Learning to teach in cooperative multiagent reinforcement learning}.
\newblock \emph{AAAI Conference on Artificial Intelligence}, 2019.

\bibitem[Garc{\'\i}a and Weiss(2019)]{garcia2019teacher}
E.~Garc{\'\i}a and E.~Weiss.
\newblock The teacher shortage is real, large and growing, and worse than we
  thought. the first report in" the perfect storm in the teacher labor market"
  series.
\newblock \emph{Economic Policy Institute}, 2019.

\bibitem[Silver et~al.(2016)Silver, Huang, Maddison, Guez, Sifre, van~den
  Driessche, Schrittwieser, Antonoglou, Panneershelvam, Lanctot, Dieleman,
  Grewe, Nham, Kalchbrenner, Sutskever, Lillicrap, Leach, Kavukcuoglu, Graepel,
  and Hassabis]{alphago}
D.~Silver, A.~Huang, C.~J. Maddison, A.~Guez, L.~Sifre, G.~van~den Driessche,
  J.~Schrittwieser, I.~Antonoglou, V.~Panneershelvam, M.~Lanctot, S.~Dieleman,
  D.~Grewe, J.~Nham, N.~Kalchbrenner, I.~Sutskever, T.~Lillicrap, M.~Leach,
  K.~Kavukcuoglu, T.~Graepel, and D.~Hassabis.
\newblock Mastering the game of go with deep neural networks and tree search.
\newblock \emph{Nature}, 529:\penalty0 484--503, 2016.

\bibitem[Coates et~al.(2009)Coates, Abbeel, and Ng]{helicopter}
A.~Coates, P.~Abbeel, and A.~Y. Ng.
\newblock Apprenticeship learning for helicopter control.
\newblock \emph{Communications of the ACM}, 52\penalty0 (7):\penalty0 97–105,
  2009.

\bibitem[Suárez-Ruiz et~al.(2018)Suárez-Ruiz, Zhou, and Pham]{ikea}
F.~Suárez-Ruiz, X.~Zhou, and Q.-C. Pham.
\newblock Can robots assemble an ikea chair?
\newblock \emph{Science Robotics}, 3\penalty0 (17), 2018.

\bibitem[Nikolaidis and Shah(2013)]{crosstraining}
S.~Nikolaidis and J.~Shah.
\newblock {Human-robot cross-training: Computational formulation, modeling and
  evaluation of a human team training strategy}.
\newblock In \emph{ACM/IEEE International Conference on Human-Robot
  Interaction}, 2013.

\bibitem[Chen et~al.(2020)Chen, Soh, Hsu, Nikolaidis, and Srinivasa]{trust}
M.~Chen, H.~Soh, D.~Hsu, S.~Nikolaidis, and S.~Srinivasa.
\newblock {Trust-aware decision making for human-robot collaboration: Model
  learning and planning}.
\newblock \emph{ACM Transactions on Human-Robot Interaction}, 9\penalty0
  (2):\penalty0 1--23, 2020.

\bibitem[Hadfield-Menell et~al.(2016)Hadfield-Menell, Russell, Abbeel, and
  Dragan]{cirl}
D.~Hadfield-Menell, S.~J. Russell, P.~Abbeel, and A.~Dragan.
\newblock Cooperative inverse reinforcement learning.
\newblock In \emph{Advances in Neural Information Processing Systems}, 2016.

\bibitem[Hambleton and Swaminathan(2013)]{irt}
R.~Hambleton and H.~Swaminathan.
\newblock \emph{Item Response Theory: Principles and Applications}.
\newblock Evaluation in education and human services. 2013.

\bibitem[Jones(2007)]{jones2007student}
L.~Jones.
\newblock \emph{The Student-centered Classroom}.
\newblock 2007.

\bibitem[Nikolaidis and Shah(2013)]{Nikolaidis2013}
S.~Nikolaidis and J.~Shah.
\newblock {Human-robot cross-training: Computational formulation, modeling and
  evaluation of a human team training strategy}.
\newblock \emph{ACM/IEEE International Conference on Human-Robot Interaction},
  2013.

\bibitem[Dragan and Srinivasa(2013)]{Dragan2013}
A.~D. Dragan and S.~S. Srinivasa.
\newblock {A policy-blending formalism for shared control}.
\newblock \emph{International Journal of Robotics Research}, 32\penalty0
  (7):\penalty0 790--805, 2013.

\bibitem[Reddy et~al.(2018)Reddy, Dragan, and Levine]{ReddyDL18}
S.~Reddy, A.~D. Dragan, and S.~Levine.
\newblock Shared autonomy via deep reinforcement learning.
\newblock In \emph{Robotics: Science and Systems}, 2018.

\bibitem[Leike et~al.(2018)Leike, Krueger, Everitt, Martic, Maini, and
  Legg]{Leike2008}
J.~Leike, D.~Krueger, T.~Everitt, M.~Martic, V.~Maini, and S.~Legg.
\newblock Scalable agent alignment via reward modeling: a research direction.
\newblock \emph{CoRR}, 2018.

\bibitem[Jeon et~al.(2020)Jeon, Milli, and Dragan]{jeon2020}
H.~J. Jeon, S.~Milli, and A.~Dragan.
\newblock Reward-rational (implicit) choice: A unifying formalism for reward
  learning.
\newblock In \emph{Advances in Neural Information Processing Systems}, 2020.

\bibitem[Christiano et~al.(2017)Christiano, Leike, Brown, Martic, Legg, and
  Amodei]{chris2017}
P.~F. Christiano, J.~Leike, T.~Brown, M.~Martic, S.~Legg, and D.~Amodei.
\newblock Deep reinforcement learning from human preferences.
\newblock In \emph{Advances in Neural Information Processing Systems}, 2017.

\bibitem[Fern et~al.(2014)Fern, Natarajan, Judah, and Tadepalli]{Fern2014}
A.~Fern, S.~Natarajan, K.~Judah, and P.~Tadepalli.
\newblock {A decision-theoretic model of assistance}.
\newblock \emph{Journal of Artificial Intelligence Research}, 50:\penalty0
  71--104, 2014.

\bibitem[Shah et~al.(2021)Shah, Freire, Alex, Freedman, Krasheninnikov, Chan,
  Dennis, Abbeel, Dragan, and Russell]{shah2021benefits}
R.~Shah, P.~Freire, N.~Alex, R.~Freedman, D.~Krasheninnikov, L.~Chan, M.~D.
  Dennis, P.~Abbeel, A.~Dragan, and S.~Russell.
\newblock Benefits of assistance over reward learning, 2021.

\bibitem[Macindoe et~al.(2021)Macindoe, Pack~Kaelbling, and
  Lozano-Pérez]{sidekick_2021}
O.~Macindoe, L.~Pack~Kaelbling, and T.~Lozano-Pérez.
\newblock Pomcop: Belief space planning for sidekicks in cooperative games.
\newblock \emph{AAAI Conference on Artificial Intelligence and Interactive
  Digital Entertainment}, 2021.

\bibitem[Grosz and Kraus(1996)]{collaborativeplan}
B.~J. Grosz and S.~Kraus.
\newblock Collaborative plans for complex group action.
\newblock \emph{Artificial Intelligence}, 86\penalty0 (2):\penalty0 269--357,
  1996.

\bibitem[Nikolaidis et~al.(2016)Nikolaidis, Kuznetsov, Hsu, and
  Srinivasa]{Nikolaidis2016}
S.~Nikolaidis, A.~Kuznetsov, D.~Hsu, and S.~Srinivasa.
\newblock Formalizing human-robot mutual adaptation via a bounded memory based
  model.
\newblock In \emph{ACM/IEEE International Conference on Human Robot
  Interaction}, 2016.

\bibitem[Nikolaidis et~al.(2017{\natexlab{a}})Nikolaidis, Hsu, and
  Srinivasa]{Nikolaidis2017}
S.~Nikolaidis, D.~Hsu, and S.~Srinivasa.
\newblock {Human-robot mutual adaptation in collaborative tasks: Models and
  experiments}.
\newblock \emph{International Journal of Robotics Research}, 36,
  2017{\natexlab{a}}.

\bibitem[Nikolaidis et~al.(2017{\natexlab{b}})Nikolaidis, Zhu, Hsu, and
  Srinivasa]{Nikolaidis2017hri}
S.~Nikolaidis, Y.~X. Zhu, D.~Hsu, and S.~Srinivasa.
\newblock Human-robot mutual adaptation in shared autonomy.
\newblock In \emph{ACM/IEEE International Conference on Human-Robot
  Interaction}, 2017{\natexlab{b}}.

\bibitem[Bengio et~al.(2009)Bengio, Louradour, Collobert, and
  Weston]{cirriculumlearning}
Y.~Bengio, J.~Louradour, R.~Collobert, and J.~Weston.
\newblock Curriculum learning.
\newblock In \emph{International Conference on Machine Learning}, 2009.

\bibitem[Zhu(2015)]{mt}
X.~Zhu.
\newblock Machine teaching: An inverse problem to machine learning and an
  approach toward optimal education.
\newblock \emph{AAAI Conference on Artificial Intelligence}, 2015.

\bibitem[Liu et~al.(2016)Liu, Zhu, and Ohannessian]{mt1}
J.~Liu, X.~Zhu, and H.~Ohannessian.
\newblock The teaching dimension of linear learners.
\newblock In \emph{International Conference on Machine Learning}, 2016.

\bibitem[Mei and Zhu(2015)]{mt2}
S.~Mei and X.~Zhu.
\newblock Using machine teaching to identify optimal training-set attacks on
  machine learners.
\newblock In \emph{AAAI Conference on Artificial Intelligence}, 2015.

\bibitem[Khan et~al.(2011)Khan, Mutlu, and Zhu]{mt3}
F.~Khan, B.~Mutlu, and J.~Zhu.
\newblock How do humans teach: On curriculum learning and teaching dimension.
\newblock In \emph{Advances in Neural Information Processing Systems}, 2011.

\bibitem[Liu et~al.(2017)Liu, Dai, Humayun, Tay, Yu, Smith, Rehg, and
  Song]{imt}
W.~Liu, B.~Dai, A.~Humayun, C.~Tay, C.~Yu, L.~B. Smith, J.~M. Rehg, and
  L.~Song.
\newblock Iterative machine teaching.
\newblock In \emph{International Conference on Machine Learning}, 2017.

\bibitem[Gur et~al.(2022)Gur, Jaques, Miao, Choi, Tiwari, Lee, and
  Faust]{Gur2022}
I.~Gur, N.~Jaques, Y.~Miao, J.~Choi, M.~Tiwari, H.~Lee, and A.~Faust.
\newblock {Environment Generation for Zero-Shot Compositional Reinforcement
  Learning}.
\newblock \emph{Advances in Neural Information Processing Systems}, 2022.

\bibitem[Portelas et~al.(2019)Portelas, Colas, Hofmann, and
  Oudeyer]{Portelas2019}
R.~Portelas, C.~Colas, K.~Hofmann, and P.-Y. Oudeyer.
\newblock {Teacher algorithms for curriculum learning of Deep RL in
  continuously parameterized environments}.
\newblock \emph{Annual Conference on Robot Learning}, 2019.

\bibitem[Fontaine* et~al.(2021)Fontaine*, Hsu*, Zhang*, Tjanaka, and
  Nikolaidis]{Fontaine*2021}
M.~Fontaine*, Y.-C. Hsu*, Y.~Zhang*, B.~Tjanaka, and S.~Nikolaidis.
\newblock {On the Importance of Environments in Human-Robot Coordination}.
\newblock \emph{Robotics: Science and Systems}, 2021.

\bibitem[Brown and Niekum(2019)]{Brown2019}
D.~S. Brown and S.~Niekum.
\newblock {Machine teaching for inverse reinforcement learning: Algorithms and
  applications}.
\newblock \emph{AAAI Conference on Artificial Intelligence}, 2019.

\bibitem[Gerkey and Matarić(2004)]{brian2004}
B.~P. Gerkey and M.~J. Matarić.
\newblock A formal analysis and taxonomy of task allocation in multi-robot
  systems.
\newblock \emph{The International Journal of Robotics Research}, 23\penalty0
  (9):\penalty0 939--954, 2004.

\bibitem[Gonz{\'a}lez-Brenes and Mostow(2013)]{GonzlezBrenes2013WhatAW}
J.~P. Gonz{\'a}lez-Brenes and J.~Mostow.
\newblock What and when do students learn? fully data-driven joint estimation
  of cognitive and student models.
\newblock In \emph{Educational Data Mining}, 2013.

\bibitem[Kim et~al.(2020)Kim, Liu, Omidshafiei, Lopez-Cot, Riemer, Habibi,
  Tesauro, Mourad, Campbell, and How]{Kim2020}
D.~K. Kim, M.~Liu, S.~Omidshafiei, S.~Lopez-Cot, M.~Riemer, G.~Habibi,
  G.~Tesauro, S.~Mourad, M.~Campbell, and J.~P. How.
\newblock {Learning hierarchical teaching policies for cooperative agents}.
\newblock \emph{International Joint Conference on Autonomous Agents and
  Multiagent Systems}, 2020.

\bibitem[Jaques et~al.(2019)Jaques, Lazaridou, Hughes, Gulcehre, Ortega,
  Strouse, Leibo, and de~Freitas]{Jaques2019}
N.~Jaques, A.~Lazaridou, E.~Hughes, C.~Gulcehre, P.~A. Ortega, D.~J. Strouse,
  J.~Z. Leibo, and N.~de~Freitas.
\newblock {Social influence as intrinsic motivation for multi-agent deep
  reinforcement learning}.
\newblock \emph{International Conference on Machine Learning}, 2019.

\bibitem[Rafferty et~al.(2011)Rafferty, Brunskill, Griffiths, and
  Shafto]{pomdpteaching}
A.~N. Rafferty, E.~Brunskill, T.~L. Griffiths, and P.~Shafto.
\newblock Faster teaching by pomdp planning.
\newblock In \emph{Artificial Intelligence in Education}, pages 280--287, 2011.

\bibitem[Singla et~al.(2014)Singla, Bogunovic, Bart{\'{o}}k, Karbasi, and
  Krause]{Singla2014}
A.~Singla, I.~Bogunovic, G.~Bart{\'{o}}k, A.~Karbasi, and A.~Krause.
\newblock {Near-optimally teaching the crowd to classify}.
\newblock \emph{International Conference on Machine Learning}, 2014.

\bibitem[Zilles et~al.(2011)Zilles, Lange, Holte, and Zinkevich]{Zilles2011}
S.~Zilles, S.~Lange, R.~Holte, and M.~Zinkevich.
\newblock {Models of cooperative teaching and learning}.
\newblock \emph{Journal of Machine Learning Research}, 12:\penalty0 349--384,
  2011.

\bibitem[Doliwa et~al.(2010)Doliwa, Simon, and Zilles]{Doliwa2010}
T.~Doliwa, H.~U. Simon, and S.~Zilles.
\newblock {Recursive teaching dimension, learning complexity, and maximum
  classes}.
\newblock \emph{Lecture Notes in Computer Science (including subseries Lecture
  Notes in Artificial Intelligence and Lecture Notes in Bioinformatics)}, 2010.

\bibitem[Mac~Aodha et~al.(2018)Mac~Aodha, Su, Chen, Perona, and
  Yue]{Aodha_2018_CVPR}
O.~Mac~Aodha, S.~Su, Y.~Chen, P.~Perona, and Y.~Yue.
\newblock Teaching categories to human learners with visual explanations.
\newblock In \emph{IEEE Conference on Computer Vision and Pattern Recognition},
  2018.

\bibitem[Srivastava et~al.(2022)Srivastava, Biyik, Mirchandani, Goodman, and
  Sadigh]{srivastava2022assistive}
M.~Srivastava, E.~Biyik, S.~Mirchandani, N.~Goodman, and D.~Sadigh.
\newblock Assistive teaching of motor control tasks to humans.
\newblock In \emph{Advances in Neural Information Processing Systems}, 2022.

\bibitem[Evrard and Kheddar(2009)]{kheddar2009}
P.~Evrard and A.~Kheddar.
\newblock Homotopy switching model for dyad haptic interaction in physical
  colåålaborative tasks.
\newblock In \emph{Joint Euro Haptics conference and Symposium on Haptic
  Interfaces for Virtual Environment and Teleoperator Systems}, 2009.

\bibitem[Kheddar(2011)]{kheddar2011}
A.~Kheddar.
\newblock Human-robot haptic joint actions is an equal control-sharing approach
  possible?
\newblock In \emph{International Conference on Human System Interactions},
  pages 268--273, 2011.

\bibitem[Graves(2012)]{Graves2012label}
A.~Graves.
\newblock \emph{{Supervised Sequence Labelling with Recurrent Neural
  Networks}}.
\newblock Studies in computational intelligence. 2012.

\bibitem[Shiarlis et~al.(2018)Shiarlis, Wulfmeier, Salter, Whiteson, and
  Posner]{shiarlis2018}
K.~Shiarlis, M.~Wulfmeier, S.~Salter, S.~Whiteson, and I.~Posner.
\newblock {TACO}: Learning task decomposition via temporal alignment for
  control.
\newblock In \emph{International Conference on Machine Learning}, 2018.

\bibitem[Kipf et~al.(2019)Kipf, Li, Dai, Zambaldi, Sanchez-Gonzalez,
  Grefenstette, Kohli, and Battaglia]{kipf2019compositional}
T.~Kipf, Y.~Li, H.~Dai, V.~Zambaldi, A.~Sanchez-Gonzalez, E.~Grefenstette,
  P.~Kohli, and P.~Battaglia.
\newblock Compile: Compositional imitation learning and execution.
\newblock In \emph{International Conference on Machine Learning}, 2019.

\bibitem[Jarrassé et~al.(2012)Jarrassé, Charalambous, and
  Burdet]{natha2012framework}
N.~Jarrassé, T.~Charalambous, and E.~Burdet.
\newblock A framework to describe, analyze and generate interactive motor
  behaviors.
\newblock \emph{Plos One}, 7\penalty0 (11):\penalty0 1--13, 11 2012.

\bibitem[Mörtl et~al.(2012)Mörtl, Lawitzky, Kucukyilmaz, Sezgin, Basdogan,
  and Hirche]{roleofrole}
A.~Mörtl, M.~Lawitzky, A.~Kucukyilmaz, M.~Sezgin, C.~Basdogan, and S.~Hirche.
\newblock The role of roles: Physical cooperation between humans and robots.
\newblock \emph{The International Journal of Robotics Research}, 31\penalty0
  (13):\penalty0 1656--1674, 2012.

\bibitem[Ekanadham and Karklin(2017)]{Ekanadham2017TSKIRTOE}
C.~Ekanadham and Y.~Karklin.
\newblock T-skirt: Online estimation of student proficiency in an adaptive
  learning system.
\newblock \emph{Machine Learning for Education Workshop at ICML}, 2017.

\bibitem[Wilson et~al.(2016)Wilson, Karklin, Han, and
  Ekanadham]{Wilson2016BackTT}
K.~H. Wilson, Y.~Karklin, B.~Han, and C.~Ekanadham.
\newblock Back to the basics: Bayesian extensions of irt outperform neural
  networks for proficiency estimation.
\newblock In \emph{Educational Data Mining}, 2016.

\bibitem[Gonzalez-Brenes et~al.(2014)Gonzalez-Brenes, Huang, and
  Brusilovsky]{pittir26017}
J.~Gonzalez-Brenes, Y.~Huang, and P.~Brusilovsky.
\newblock General features in knowledge tracing to model multiple subskills,
  temporal item response theory, and expert knowledge.
\newblock In \emph{Educational Data Mining}, 2014.

\bibitem[Lupu et~al.(2021)Lupu, Cui, Hu, and Foerster]{lupu21a}
A.~Lupu, B.~Cui, H.~Hu, and J.~Foerster.
\newblock Trajectory diversity for zero-shot coordination.
\newblock In \emph{International Conference on Machine Learning}, 2021.

\bibitem[Zhao et~al.(2021)Zhao, Song, Haifeng, Gao, Wu, Sun, and Wei]{Zhao2021}
R.~Zhao, J.~Song, H.~Haifeng, Y.~Gao, Y.~Wu, Z.~Sun, and Y.~Wei.
\newblock {Maximum Entropy Population Based Training for Zero-Shot Human-AI
  Coordination}.
\newblock \emph{CoRR}, 2021.

\bibitem[Carroll et~al.(2019)Carroll, Shah, Ho, Griffiths, Seshia, Abbeel, and
  Dragan]{Carroll2019OnTU}
M.~Carroll, R.~Shah, M.~K. Ho, T.~L. Griffiths, S.~A. Seshia, P.~Abbeel, and
  A.~D. Dragan.
\newblock On the utility of learning about humans for human-ai coordination.
\newblock In \emph{Advances in Neural Information Processing Systems}, 2019.

\bibitem[Knott et~al.(2021)Knott, Carroll, Devlin, Ciosek, Hofmann, Dragan, and
  Shah]{knott2021eval}
P.~Knott, M.~Carroll, S.~Devlin, K.~Ciosek, K.~Hofmann, A.~D. Dragan, and
  R.~Shah.
\newblock Evaluating the robustness of collaborative agents.
\newblock In \emph{International Joint Conference on Autonomous Agents and
  Multiagent Systems}, 2021.

\bibitem[Charakorn et~al.(2020)Charakorn, Manoonpong, and
  Dilokthanakul]{Charakorn2020InvestigatingPD}
R.~Charakorn, P.~Manoonpong, and N.~Dilokthanakul.
\newblock Investigating partner diversification methods in cooperative
  multi-agent deep reinforcement learning.
\newblock In \emph{Iconip}, 2020.

\bibitem[Nalepka et~al.(2021)Nalepka, Gregory-Dunsmore, Simpson, Patil, and
  Richardson]{nal2021intention}
P.~Nalepka, J.~Gregory-Dunsmore, J.~Simpson, G.~Patil, and M.~Richardson.
\newblock Interaction flexibility in artificial agents teaming with humans.
\newblock In \emph{Annual Meeting of the Cognitive Science Society}, 2021.

\bibitem[Sarkar et~al.(2022)Sarkar, Talati, Shih, and
  Dorsa]{sarkar2021pantheonRL}
B.~Sarkar, A.~Talati, A.~Shih, and S.~Dorsa.
\newblock Pantheonrl: A marl library for dynamic training interactions.
\newblock In \emph{AAAI Conference on Artificial Intelligence (Demo Track)},
  2022.

\bibitem[Wang et~al.(2021)Wang, Gupta, Mahajan, Peng, Whiteson, and
  Zhang]{wang2021rode}
T.~Wang, T.~Gupta, A.~Mahajan, B.~Peng, S.~Whiteson, and C.~Zhang.
\newblock {\{}RODE{\}}: Learning roles to decompose multi-agent tasks.
\newblock In \emph{International Conference on Learning Representations}, 2021.

\end{thebibliography}
